\documentclass[11pt]{article}

\usepackage[preprint]{acl}

\usepackage{times}
\usepackage{latexsym}
\usepackage{comment}
\usepackage{booktabs}
\usepackage{array}
\usepackage{multirow}
\usepackage{amsmath}
\usepackage{enumitem}
\usepackage{arydshln}
\usepackage{float}
\usepackage[subtle]{savetrees}
\usepackage[T1]{fontenc}

\usepackage[utf8]{inputenc}

\usepackage{microtype}

\usepackage{inconsolata}

\usepackage{graphicx}

\newcommand{\method}{DTAD}
\newcommand{\vicl}{vICL}
\newcommand{\data}{MIME}

\title{Watch and Learn: Teaching Visual In-Context Learning \\ by Removing Textual Shortcuts}

\author{
  Nimrod Shabtay$^{1,2}$ \quad
  Asaf Yehudai$^{1}$ \quad
  Avihu Dekel$^{1}$ \quad
  Eli Schwartz$^{1}$ \quad
  Raja Giryes$^{2}$ \\
  $^{1}$IBM Research \quad
  $^{2}$Tel-Aviv University
}

\begin{document}
\maketitle

\begin{figure*}[t]
\centering
\includegraphics[width=0.8\linewidth]{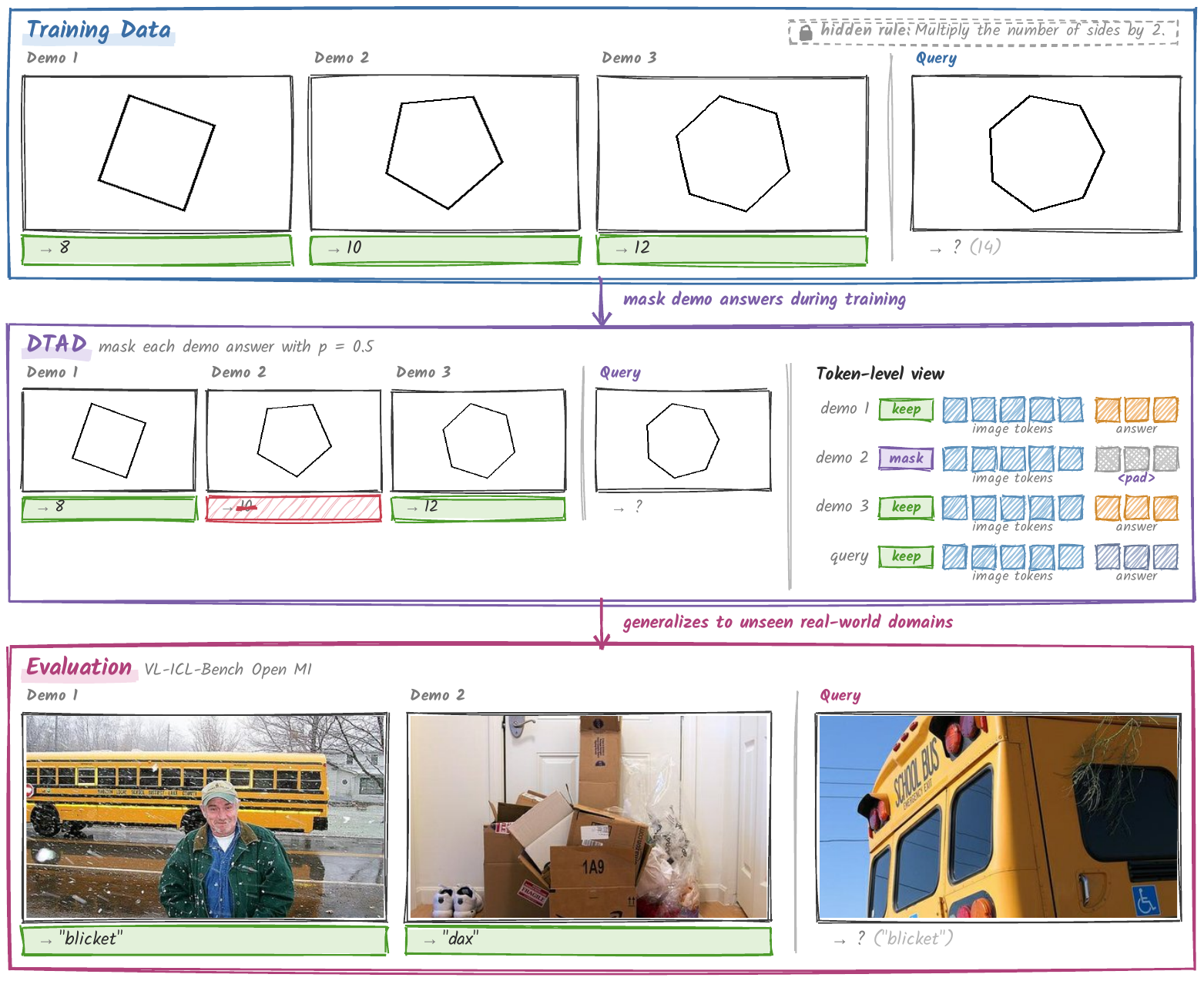} 
\caption{We fine-tune on \data{}, synthetic in-context instances whose task is a hidden rule that must be inferred from the demonstrations. Despite being synthetic, \data{} improves results on unseen real-world benchmarks. \method{}, which masks demonstration answers, produces more stable attention patterns that focus on the demonstration images.}
\label{fig:teaser}
\vspace{-10pt}
\end{figure*}

\begin{abstract}
Vision-language models (VLMs) struggle with in-context learning compared to their textual counterparts. Our preliminary study shows that VLMs rely on textual shortcuts and underutilize the visual demonstrations. We therefore hypothesize that VLMs are insufficiently trained to perform visual in-context learning (\vicl{}).
To this end, we construct \data{}, a task-agnostic \vicl{} training set without textual shortcuts. We ensure the instruction does not state the rule linking the demonstration images to their answers, so the model must infer it from the visual demonstrations.
Fine-tuning on \data{} improves \vicl{} accuracy by 8.1--18.0 points across three VLMs, preserves single-image performance, and improves multi-image and text-only in-context learning. In contrast, stating the rule in the training data sharply reduces this gain. We further propose \method{} (Demonstration Text Attention Dropout), which randomly withholds demonstration labels during training so that the VLM must infer the rule from the demonstration images.
While achieving comparable \vicl{} accuracy to fine-tuning on \data{}, \method{} yields more consistent attention to the demonstration images.
On TrueMICL, Qwen3-VL-8B trained on \data{} matches frontier models such as Claude Opus 5, and surpasses them with \method{}.
Finally, adding \data{} to the instruction-tuning mix of LLaVA-1.5-13B more than doubles its \vicl{} accuracy without hurting single-image abilities, whereas a same-size multi-image set improves it only slightly. These results suggest that \vicl{} data should be a standard component of VLM training.
\end{abstract}

\section{Introduction}
\label{sec:intro}
In-context learning is a defining emergent capability of large language models (LLMs): given a handful of demonstrations, an LLM infers the latent task and applies it to a new query without any parameter update \citep{brown2020language,todd2024function,olsson2022context}. Vision-language models (VLMs) are built on LLMs and are routinely prompted with few-shot demonstrations, yet they struggle with visual in-context learning (\vicl{}), where the task must be inferred from the demonstration images rather than from the accompanying text \citep{baldassini2024makes, chen2025can, santos2025vision, wang2026multimodal, jia2025symdpo}.
One plausible cause is the pretraining data. In text, the demonstration-query-answer structure occurs naturally at pretraining scale \citep{han2023understanding,hahn2023theory,chen2024parallel}, whereas its multimodal analogue is likely absent \citep{doveh2024towards}.

Prior work on improving \vicl{} follows three approaches. The first is motivated by the finding that VLMs attend to the demonstration text rather than the demonstration images \citep{chen2025true, jia2025symdpo}, and rebalances attention toward the images. \citet{chen2025true} train an attention-reweighting layer (DARA), while \citet{wang2026multimodal} reweight attention at inference time (MGI). The second extracts task-specific steering or task vectors from many-shot activations \citep{huang2024multimodal, li2025m2iv}. The third fits demonstration-retrieval policies per task \citep{zhang2023makes, doveh2024towards, doveh2025teaching}. All three are tailored to a specific task or model, and none of them yields a general \vicl{} ability.

We start by asking why VLMs struggle with \vicl{} (\S\ref{subsec:prelim}). On two TrueMICL tasks, Qwen3-VL-8B reaches near-perfect accuracy (99.8\%) when the demonstrations are given as text, or when it first transcribes the images itself. Hence, neither perception nor core in-context learning is the bottleneck. Instead, the model relies on textual shortcuts. Its accuracy falls from 78.3\% to 44.8\% when the task-specific instruction is replaced with a generic one that gives no hint about the task. The model thus underutilizes the visual demonstrations, even when the task can only be learned from them.
We therefore hypothesize that VLMs are insufficiently trained to perform \vicl{}, and that training on dedicated \vicl{} data without textual shortcuts will teach them to attend to and learn from the demonstration images.

Accordingly, we curate \data{} (Multi-source Induction from Mute Episodes), a diverse, task-agnostic \vicl{} training set (\S\ref{subsec::data}). It spans five sources with different rule types and visual domains, so the model cannot solve it by learning a single task. We construct each instance so that no text states the rule linking the demonstration images to their answers, and the model can infer it only from the demonstrations.
To further shift attention from the demonstration text to the demonstration images, we propose \method{} (Demonstration Text Attention Dropout, \S\ref{subsec::method_name}). During fine-tuning, \method{} randomly withholds each demonstration's label. A demonstration without its label can inform the query only through its image, so the model must learn to read the demonstration images.

In our experiments, fine-tuning on \data{} improves average \vicl{} accuracy by 8.1--18.0 points across 11 tasks and three VLMs, whereas DARA and MGI leave the base model unchanged (\S\ref{subsec:res_data}).
Training on \data{} also preserves single-image performance and improves multi-image and text-only in-context learning (\S\ref{subsec:res_beyond}). This suggests that the model learns a general way of reading demonstrations rather than an image-specific patch. \method{} matches the accuracy of fine-tuning on \data{} while attending to the demonstration images more strongly and more consistently across tasks and shot counts (\S\ref{subsec:visual_pathway}). It is also more robust to changes in the prompt format (\S\ref{app:format_ablation}). On TrueMICL, Qwen3-VL-8B trained on \data{} matches frontier models such as Claude Opus 5, and surpasses them with \method{}.

Our analysis attributes these gains to the design of \data{} (\S\ref{sec:analysis}). Models trained on \data{} depend on their demonstrations. Shuffling either the images or the labels hurts them more than the base model, showing that they learn the rule from the image-label pairs (\S\ref{sec:demo_reliance}). Hiding the rule is necessary. Training on the same instances with the rule stated sharply reduces the gain, and the model then attends to the demonstration images less than the base model in the mid-to-late layers (\S\ref{sec:rule_hidden}).

\data{} also belongs in instruction tuning, not only in post-hoc fine-tuning. We retrain LLaVA-1.5-13B \citep{liu2023llava} on its original instruction-tuning mix, adding either \data{} or a same-size set of multi-image conversations (\S\ref{subsec:res_retrain}). \data{} more than doubles TrueMICL accuracy (18.9 to 39.7) with no single-image regression, whereas the multi-image data barely helps (21.5) and lowers POPE by 30 points. These results suggest that \vicl{} data should be a standard component of VLM training.





\section{Preliminary Study: The Text Shortcut in \vicl{}}
\label{subsec:prelim}

We investigate the source of VLMs' shortcomings in \vicl{} through a preliminary study using Qwen3-VL-8B~\cite{bai2025qwen3} on two tasks from TrueMICL~\cite{chen2025true}. We consider four alternative settings that progressively remove textual cues to examine whether the model can learn from visual demonstrations independently of textual shortcuts. The first uses task metadata to formulate the task as pure text-ICL. The second transcribes the images into text before answering the question based on the resulting descriptions. The third uses a per-task instruction (\texttt{``What is the mathematical result of the blue number and the red number?''})
and the fourth uses a generic instruction (\texttt{``Look at the demonstrations and answer the final query.''}) that provides no task-specific hint.

Table~\ref{tab:app_prelim} reports the results. Both pure text-ICL and the transcribe-and-answer settings achieve near-perfect accuracy ($\sim{}$100\%), indicating that the model can both perceive the relevant visual information and perform ICL over textual demonstrations. This suggests that neither visual perception nor textual ICL alone explains the performance gap in \vicl{}. The task-specific setting achieves 78.3\% accuracy, and removing the task-specific hint further reduces accuracy to 44.8\%, suggesting that the model relies heavily on textual cues rather than learning from the visual demonstrations themselves.

These findings motivate our hypothesis that current VLMs are insufficiently trained to perform ICL over visual demonstrations. This limitation may also reflect a tendency to underutilize visual tokens when solving tasks that require learning from contextual examples. If so, fine-tuning on vICL data designed to eliminate textual shortcuts could encourage models to attend to and learn from visual contextual signals.




\section{Enhancing General \vicl{} abilities}
\vspace{-4pt}
\label{sec:method}
\subsection{\data{}: A Task-Agnostic \vicl{} Training Set}
\label{subsec::data}
We build \data{} with instances where the rule is deducible only by using the demonstration images. Hence, a model can solve an instance only by inducing the rule across shots.
\data{} contains 23{,}578 instances (Table~\ref{tab:dataset_composition}). Each instance has 1--5 demonstration shots, each an (image, label) pair, followed by a query image whose label is the held-out target. We construct \data{} to adhere to the following properties, requiring true visual induction.

\begin{enumerate}[nosep]
    \item \textbf{The rule is expressed only through the demonstrations.} The image--label pairs carry the rule, and nothing else in the instance conveys it. Shuffling either the images or the labels sharply degrades a trained model's accuracy (\S\ref{sec:demo_reliance}), so neither half carries the rule alone.
    \item \textbf{The rule is never stated.} All instances share one generic rule-agnostic instruction, so the prompt never describes the rule. A stated rule would let the model solve the instance from text alone, which is the bypass we aim to remove.
    \item \textbf{The images are needed.} Without the images, a text-only Qwen3-8B~\cite{yang2025qwen3} stays at near random chance on average.     
    \item \textbf{Demonstrations and queries are instance-disjoint.} No query repeats a demonstration image or its underlying instance, so the model cannot answer by matching the query to a demonstration.    
\end{enumerate}

We apply the schema to five diverse sources. These sources have different types of rules and different visual domains. This ensures that the model does not learn a task-specific ability:
\begin{itemize}[nosep]
    \item \textbf{Polygon}~\citep{chia2024puzzlevqa}: rules for the number of sides of a polygon.
    \item \textbf{Bongard-style}~\citep{nie2020bongard}: concepts about shape properties, for example, fill, size, and rotation.
    \item \textbf{ARC-style}~\citep{chollet2019measure, hodel2024addressing}: transforms from an input grid to an output grid.
    \item \textbf{Triangle arithmetic}~\citep{chia2024puzzlevqa}: arithmetic with the values at the vertices. We mask the value of one vertex.
    \item \textbf{CLEVR}~\citep{johnson2017clevr}: counting and arithmetic with the objects in a frame.
\end{itemize}

We ablate these design choices in \S\ref{sec:rule_hidden} and Appendix~\ref{app:source_contribution}, and verify the properties in \S\ref{sec:data_analysis} and Appendix~\ref{app:design_choices}.

\subsection{\method{}: Demonstration Text Attention Dropout}
\label{subsec::method_name}

\data{} already pushes the model to extract the task from the demonstration images. \method{} strengthens this by targeting the text-over-image bias directly, as VLMs attend to the demonstration text rather than the demonstration images (Figure~\ref{fig:teaser}). For each demonstration,
we independently withhold its text label with probability $p$ and keep its image. A withheld label leaves only one path from that demonstration to the query label, through its image, so minimizing the query loss requires reading the image. Because demonstrations are dropped independently, an instance mixes fully visible and image-only demonstrations, and the model learns to use both routes jointly. 
The query is always shown in full and is the sole target of the loss. Inference is unchanged.

We withhold labels in one of two ways: replacing the label tokens with a placeholder, or blocking attention to them at every layer. Both preserve the conversation structure, sequence length, and image content. We compare \method{} against fine-tuning on the same data in \S\ref{subsec:res_dtad}.

\section{Experimental Results}
\label{sec:results}
 
We organize the results around four findings. First, fine-tuning on \data{} improves \vicl{} on all three backbones, while existing attention-rebalancing methods leave the baseline unchanged. With \data{}, an 8B model matches frontier models, and surpasses them with \method{} (\S\ref{subsec:res_data}). Second, the gain comes from the demonstration structure of \data{}, not from multi-image exposure. Added to LLaVA-1.5's full instruction-tuning mix, it more than doubles \vicl{} accuracy with no single-image regression (\S\ref{subsec:res_retrain}). Third, \method{} matches this accuracy while attending to the demonstration images more stably (\S\ref{subsec:res_dtad}). Finally, training on \data{} preserves single-image performance and transfers to multi-image, abstract visual reasoning, and text-only ICL (\S\ref{subsec:res_beyond}).
 
\subsection{Experimental Setup}
\label{subsec:res_setup}
 
\noindent\textbf{Models.} We use three models: Qwen3-VL-8B~\citep{bai2025qwen3}, LLaVA-OneVision-1.5-8B~\citep{an2025llava}, and Granite-Vision-4.1-4B~\citep{team2025granite}. We fine-tune each model with LoRA ($r{=}8$) for one epoch on \data{}, with a learning rate of $2\times10^{-4}$ and an effective batch size of 16. For \method{}, we use $p{=}0.5$ with token replacement (TR). We report the attention-mask (AM) variant as an ablation (\S\ref{subsec:visual_pathway}). 
 
\noindent\textbf{Benchmarks.} We evaluate on 11 tasks: Five TrueMICL tasks~\citep{chen2025true} (Clock, Operator Induction, Palindrome, Shapes, Classification), the vICL task of SEED-Bench-2 (Task~23)~\citep{li2024seed}, and five VL-ICL-Bench tasks~\citep{zong2024vlicl} (Operator Induction, Interleaved, TextOCR, Open MI, Match MI). We exclude the Sudoku task, where every model has a perfect score, and CLEVR-based benchmarks, whose images overlap the training data (\S\ref{sec:data_analysis}). We report accuracy averaged over 1/2/4-shots and three demonstration seeds.
 
\noindent\textbf{Two protocols.} The \emph{official} protocol keeps each benchmark's task-specific question. The \emph{generic instruction} protocol replaces it with one generic instruction for all tasks (\texttt{``Each demonstration shows an image and the correct answer. Look at the demonstrations and answer the final query.''}), so the rule can only be deduced from the demonstrations. We adopt the generic instruction protocol as our primary protocol and use it henceforth throughout the paper 
The results under the official protocol are reported in Appendix~\ref{app:official}.
 
\noindent\textbf{Baselines.} We compare our approach against the untrained Qwen3-VL models, DARA~\citep{chen2025true}, and inference-time attention rebalancing (MGI)~\citep{wang2026multimodal}. We use the official Qwen3-VL implementations of DARA and MGI.

\begin{table*}[t]
\centering
\small
\setlength{\tabcolsep}{4pt}
\caption{\textbf{Visual ICL results (generic instruction protocol).} Accuracy (\%) of each base model before and after fine-tuning on \data{}, averaged over 1/2/4 shots and three seeds.}
\label{tab:main}
\vspace{-6pt}
\resizebox{\textwidth}{!}{%
\begin{tabular}{l ccccc c ccccc cc}
\toprule
 & \multicolumn{5}{c}{TrueMICL} & SB-2 & \multicolumn{5}{c}{VL-ICL-Bench} & & \\
\cmidrule(lr){2-6}\cmidrule(lr){7-7}\cmidrule(lr){8-12}
Model & Clock & Op.Ind & Palin & Shapes & Classif & T23 & Op.Ind & Interl & TextOCR & Open MI & Match MI & Avg & $\Delta$ \\
\midrule
Qwen3-VL-8B        & 12.8 & 76.7 & 63.9 & 95.8 & 45.7 & 35.8 & 78.5 & 59.4 & 59.2 & 72.1 & 85.8 & 62.3$_{\pm0.9}$ & -- \\
\quad + DARA~\citep{chen2025true}                & 12.6 & 76.7 & 64.2 & 95.8 & 45.3 & 35.8 & 78.3 & 59.8 & 59.5 & 71.8 & 85.8 & 62.3 & +0.0 \\
\quad + MGI~\citep{wang2026multimodal}                 & 12.9 & 76.6 & 64.3 & 96.0 & 45.4 & 35.8 & 78.3 & 59.4 & 59.6 & 72.2 & 85.2 & 62.3 & +0.0 \\
\cdashline{1-14}
\quad + \data{}       & 62.8 & \textbf{88.9} & 68.8 & 97.3 & \textbf{63.4} & \textbf{50.0} & \textbf{91.1} & 86.9 & 58.7 & \textbf{98.2} & 85.3 & \textbf{77.4}$_{\pm1.2}$ & +15.1 \\
\quad + \data{} + \method{}    & \textbf{70.1} & 88.1 & \textbf{79.3} & \textbf{99.2} & 53.9 & 45.8 & 90.0 & \textbf{87.8} & \textbf{60.7} & 94.9 & 81.5 & \textbf{77.4}$_{\pm1.5}$ & +15.1 \\
\midrule 
LLaVA-OV-1.5-8B    & 17.8 & 74.9 & 67.1 & 92.6 & 38.9 & \textbf{41.7} & 83.1 & 56.9 & 49.5 & 51.7 & 54.7 & 57.2$_{\pm0.1}$ & -- \\
\quad + \data{}       & 30.8 & \textbf{83.6} & 63.7 & \textbf{99.6} & \textbf{55.8} & 40.8 & \textbf{88.3} & 53.7 & \textbf{56.5} & \textbf{84.4} & \textbf{60.8} & \textbf{65.3}$_{\pm0.9}$ & +8.1 \\
\quad + \data{} + \method{}    & \textbf{31.6} & 80.3 & \textbf{72.7} & 98.8 & 53.5 & 34.2 & 87.2 & \textbf{61.3} & 55.5 & 72.3 & 57.5 & 64.1$_{\pm1.3}$ & +6.9 \\
\midrule
Granite-Vision-4.1-4B & 4.0 & 41.1 & 61.2 & 53.4 & 29.8 & \textbf{30.8} & 45.6 & 9.8 & 39.8 & 49.2 & 50.0 & 37.7$_{\pm0.2}$ & -- \\
\quad + \data{}       & \textbf{24.4} & \textbf{66.8} & 62.8 & \textbf{99.0} & 43.9 & \textbf{30.8} & \textbf{73.3} & 22.6 & 39.1 & \textbf{97.0} & \textbf{53.4} & 55.7$_{\pm0.3}$ & +18.0 \\
\quad + \data{} + \method{}    & 21.0 & 57.6 & \textbf{92.6} & 98.8 & \textbf{47.4} & 29.2 & 66.5 & \textbf{37.4} & \textbf{40.0} & 84.6 & 50.1 & \textbf{56.8}$_{\pm1.5}$ & +19.1 \\
\bottomrule
\end{tabular}}
\end{table*}

\subsection{\data{} Drives the Accuracy Gain}
\label{subsec:res_data}
 
\noindent\textbf{Main result.} Table~\ref{tab:main} reports the generic instruction results. Fine-tuning on \data{} improves all three backbones, by +15.1, +8.1, and +18.0 points on average. Under the official protocol, the gains are +7.8, +9.7, and +31.6 points (Appendix Table~\ref{tab:app_official}). \method{} trained on the same data scores on par with the fine-tuned models, staying within about a one-point range on average. This indicates that the core of the improvement comes from the data itself.

\noindent\textbf{Existing methods barely change the base model's performance.} On Qwen3-VL-8B, DARA and MGI stay within $\pm$0.2 points of the untrained model under both protocols (official 69.8/69.7, generic instruction 62.3/62.3). They change only a small fraction of outputs.
 
\noindent\textbf{Outperforming frontier models on \vicl{}.} Training Qwen3-VL-8B on \data{} makes it competitive with frontier models and with open models approximately 30$\times$ larger. It outscores Qwen3-VL-235B-A22B, InternVL3.5-241B-A28B, and Claude Opus 4.6, and matches Claude Opus 5~\cite{opus46,opus5}. With \method{}, it outperforms all other models (Table~\ref{tab:app_large_models}). The advantage of \method{} is largest on tasks where the model must understand the rule from the images, and not the text.
This indicates that even frontier models do not necessarily obtain strong \vicl{} capabilities. Furthermore, this emphasizes the relevance of our approach for frontier model training.

\begin{table}[t]
\centering
\footnotesize
\caption{\textbf{Comparison to frontier and Large models on TrueMICL.} Per-task and overall accuracy on the five TrueMICL tasks. Fine-tuned on \data{}, Qwen3-VL-8B outperforms models roughly 30$\times$ larger and Claude Opus 4.6, and matches Claude Opus 5; with \method{}, it outperforms all of them.}
\label{tab:app_large_models}
\vspace{-6pt}
\resizebox{\ifdim\width>\linewidth\linewidth\else\width\fi}{!}{%
\begin{tabular}{ll|ccccc|c}
\toprule
Model & Size & Clock & Op.Ind & Palin & Shapes & Classif & Overall \\
\midrule
Qwen3-VL-235B-A22B              & 235B & 31.0 & 99.3 & 86.0 & 97.8 & \textbf{64.2} & 75.7 \\
InternVL3.5-241B-A28B           & 241B & 15.3 & \textbf{99.7} & 79.0 & 83.9 & 48.4 & 65.3 \\
\midrule
Claude Opus 4.6                 & -    & 43.3 & 90.2 & 76.7 & 99.1 & 39.6 & 69.8 \\
Claude Opus 5                   & -    & 36.1 & 88.6 & \textbf{95.1} & \textbf{99.6} & 61.0 & 76.1 \\
\midrule
Qwen3-VL-8B                     & 8B   & 12.8 & 76.7 & 63.9 & 95.8 & 45.7 & 59.0 \\
\quad + \data{}                    & 8B   & 62.8 & 88.9 & 68.8 & 97.3 & 63.4 & 76.2 \\
\quad + \data{} + \method{} & 8B & \textbf{70.1} & 88.1 & 79.3 & 99.2 & 53.9 & \textbf{78.1} \\
\bottomrule
\end{tabular}}
\vspace{-4pt}
\end{table}

\subsection{The Data Belongs in the Training Mix}
\label{subsec:res_retrain}
 
\begin{table}[t]
\centering
\small
\caption{\textbf{Full retraining of LLaVA-1.5-13B.} \data{} and a same-size multi-image set (Mantis) are each added to the original instruction-tuning mix. TrueMICL accuracy is averaged over 1/2/4 shots.}
\label{tab:retrain}
\vspace{-6pt}
\resizebox{\columnwidth}{!}{%
\begin{tabular}{l c ccc}
\toprule
 & TrueMICL & \multicolumn{3}{c}{Regression check} \\
\cmidrule(lr){3-5}
Training mix & Avg & POPE & MME-P & MME-C \\
\midrule
LLaVA-1.5-13B (official) & 18.9 & 84.2 & 1561 & 118 \\
\quad + Multi-Image data           & 21.5 & 54.1 & 1509 & 116 \\
\quad + \data{}         & \textbf{39.7} & \textbf{85.2} & \textbf{1575} & \textbf{127} \\
\bottomrule
\end{tabular}}
\vspace{-10pt}
\end{table}

To assess \data{} usefulness beyond general fine-tuning, we retrain LLaVA-1.5-13B~\citep{liu2023llava} on its original 665K instruction-tuning mix with \data{} added, and compare against adding a same-size set of multi-image conversational data (sampled from Mantis~\cite{Jiang2024mantis} training set) (Table~\ref{tab:retrain}). \data{} results more than double TrueMICL accuracy (18.9 to 39.7), while the multi-image data only improve slightly, to 21.5. Moreover, \data{} causes no regression on single-image abilities (POPE~\cite{POPE} or MME~\cite{fu2026mme}), whereas the Multi-Image training set drops POPE by 30 points. Per-task results are in Appendix Table~\ref{tab:llava_training}.
Therefore, multi-image exposure alone does not directly improve \vicl{}; the demonstration structure of \data{} does, and it can be added directly to a model's instruction training mix.

\subsection{\method{}: Improved Visual Pathway at No Accuracy Cost}
\label{subsec:res_dtad}
 
Table \ref{tab:main} shows that \method{} performs on par with fine-tuning on \data{} (+ \data{}) across the three models Qwen3-VL-8B, LLaVA-OV-1.5-8B, and Granite-Vision-4.1-4B (tied at 77.4, 64.1 vs.\ 65.3, and 56.8 vs.\ 55.7).

\noindent\textbf{Stable visual attention and robustness to prompt format.} In our attention analysis (\S\ref{subsec:visual_pathway}) we find that although \method{} drops labels and discards information, it teaches the model to attend to the demonstration images more strongly and more consistently across tasks. \method{} builds a stable visual pathway: its late layers attend to the visual signal more than the untrained model at every layer and shot count.
Similarly, \method{} is more robust to prompts. When the prompt departs from the training format, it maintains higher performance, reducing the average drop by about two-thirds relative to fine-tuning (\S\ref{app:format_ablation}).

\subsection{Beyond \vicl{} Benchmarks}
\label{subsec:res_beyond}
\begin{table}[t]
\centering
\small
\caption{\textbf{Beyond visual ICL.} SI (single-image): average of MMStar, RealWorldQA, POPE, and InfoVQA. MI (multi-image): average of Mantis and LLaVA-Interleave. MARVEL is evaluated as multi-turn visual ICL (chance 25\%). Text-only ICL uses multi-turn demonstrations.}
\label{tab:beyond}
\setlength{\tabcolsep}{3pt}
\resizebox{\columnwidth}{!}{%
\begin{tabular}{l cc c ccc}
\toprule
 & \multicolumn{2}{c}{Regression} & & \multicolumn{3}{c}{Text-only ICL} \\
\cmidrule(lr){2-3}\cmidrule(lr){5-7}
Model & SI & MI & MARVEL & MMLU & GSM8K & GPQA \\
\midrule
Qwen3-VL-8B              & 76.1 & 54.9 & 27.0 & 74.6 & 89.8 & 36.4 \\
\quad + \data{}             & 76.4 & 57.5 & 29.1 & 75.6 & 91.4 & \textbf{37.1} \\
\quad + \data{} + \method{} & \textbf{76.9} & \textbf{58.1} & \textbf{31.6} & \textbf{75.8} & \textbf{92.7} & \textbf{37.1} \\
\bottomrule
\end{tabular}%
}
\vspace{-15pt}
\end{table}
 
In Table~\ref{tab:beyond} we examine the effect of training on other abilities, assessing whether it degrades or transfers to them (Per-benchmark results in Appendix~\ref{app:beyond}).
Single-image performance is preserved (MMStar, MME, RealWorldQA, POPE, InfoVQA), and multi-image performance improves by 2.6 points (Mantis~\citep{Jiang2024mantis}, LLaVA-Interleave). On MARVEL~\citep{jiang2024marvel}, where the rule must be induced from abstract visual puzzles, visual-ICL accuracy improves from 27.0 to 29.1. Finally, the skill transfers to text-only in-context learning, improving accuracy on MMLU~\citep{hendrycks2020mmlu}, GSM8K~\citep{cobbe2021gsm8k}, and GPQA~\citep{rein2023gpqa} although training uses only visual demonstrations. This suggests the model learns a general way of reading demonstrations rather than a patch specific to images.

\section{Ablations \& Analysis}
\label{sec:analysis}

This section investigates the causes of the gains in \S\ref{sec:results}. We first show that trained models require their demonstrations and use both the images and the answers (\S\ref{sec:demo_reliance}). We then analyze the training data. We show that the design choice not to state the rule in \data{} is crucial, and stating it reduces the gain dramatically (\S\ref{sec:rule_hidden}). We also ensure the evaluation benchmarks are visually far from the training data, so the gains are not due to memorization (\S\ref{sec:data_analysis}).
Next, we demonstrate that \method{} produces more stable visual attention patterns. It uses visual pathways that are consistent across shots. We also show that when the demonstration text is absent, the model moves its attention to the demonstration images (\S\ref{subsec:visual_pathway}). We establish that \method{} is more robust to changes in the prompt format (\S\ref{app:format_ablation}).
Finally, we measure how the accuracy changes when the number of shots increases (\S\ref{sec:shot_scaling}).

\subsection{Trained models rely on the demonstrations}
\label{sec:demo_reliance}

If a model induces the rule from the demonstrations, corrupting them should hurt it. We corrupt the demonstrations in four ways: shuffling answers across demonstrations, shuffling images across demonstrations, replacing them with demonstrations of a different rule (\emph{foreign rule}), and removing them altogether.

\begin{table}[t]
\centering
\small
\setlength{\tabcolsep}{4pt}
\caption{\textbf{Trained models depend on the demonstrations.} Accuracy with correct demonstrations, and change in points under each corruption, on the five TrueMICL tasks (generic instruction protocol, three seeds).}
\label{tab:corrupt}
\vspace{-6pt}
\resizebox{\columnwidth}{!}{%
\begin{tabular}{lccccc}
\toprule
\multirow{2}{*}{Model} & \multirow{2}{*}{Correct} & Answers & Images & Foreign & No \\
 & & shuffled & shuffled & rule & demos \\
\midrule
\multicolumn{6}{l}{\textit{2 shots}} \\
Qwen3-VL-8B           & 57.5 & $-$19.3 & $-$19.8 & $-$43.5 & $-$25.3 \\
+ \data{}             & 74.5 & $-$35.7 & $-$35.1 & $-$63.8 & $-$48.7 \\
+ \method{} (AM)      & 74.3 & $-$32.5 & $-$33.4 & $-$70.7 & $-$42.3 \\
+ \method{} (TR)      & 77.8 & $-$34.6 & $-$33.6 & $-$71.4 & $-$47.6 \\
\midrule
\multicolumn{6}{l}{\textit{4 shots}} \\
Qwen3-VL-8B           & 66.6 & $-$30.9 & $-$30.5 & $-$50.2 & $-$34.2 \\
+ \data{}             & 88.6 & $-$52.3 & $-$52.0 & $-$78.6 & $-$62.9 \\
+ \method{} (AM)      & 87.7 & $-$49.5 & $-$49.3 & $-$83.6 & $-$55.7 \\
+ \method{} (TR)      & 90.0 & $-$50.7 & $-$49.3 & $-$81.5 & $-$59.9 \\
\bottomrule
\end{tabular}}
\vspace{-5pt}
\end{table}

Shuffling the answers or the images reduces the accuracy of trained models by 33--52 points, and removing the demonstrations reduces it by 42--63 points; the untrained model loses much less (Table~\ref{tab:corrupt}). The two shuffles cause about the same drop, so the rule is in the image--label pairs, not in either half alone. Foreign-rule demonstrations cause the largest drop for \method{}, larger than for fine-tuning: a \method{} model follows the rule the demonstrations show, even when that rule is wrong for the query.
\subsection{Hiding the rule is necessary}
\label{sec:rule_hidden}

We construct \data{} to never explicitly state the rule (\S\ref{subsec::data}). To test this choice effect, we train the same model on the same instances, demonstrations, and order, with one change: the prompt also states the demonstration rule. Both models use fine-tuning with identical hyperparameters.

\begin{table}[t]
\centering
\small
\caption{\textbf{Stating the rule substantially reduces the transfer gain.} Accuracy (2 / 4 shots) on the TrueMICL tasks of Table~\ref{tab:corrupt}, and change under shuffled demonstration answers.}
\label{tab:rule_stated}
\vspace{-6pt}
\begin{tabular}{lcc}
\toprule
Training data & TrueMICL & Answers shuffled \\
\midrule
Untrained & 57.5 / 66.6 & $-$19.3 / $-$30.9 \\
Rule stated & 69.6 / 78.9 & $-$26.3 / $-$42.8 \\
Rule hidden (ours) & \textbf{74.5 / 88.6} & $-$35.7 / $-$52.3 \\
\bottomrule
\end{tabular}
\vspace{-15pt}
\end{table}

Table~\ref{tab:rule_stated} shows that stating the rule reduces the gain to +12.1 / +12.3 points over the untrained model, relative to +17.0 / +22.0 when the rule is hidden. The rule-stated model is also less sensitive to shuffled answers. Given the rule, it learns to read the rule and relies less on the demonstrations than the \data-trained model. Similarly, its visual share falls below the base model in the mid-to-late layers (Figure \ref{fig:visual_heads}).

\paragraph{Single-image control.} To control for the presence of the demonstrations, we train on the same query images, with each rule stated in text, but without the demonstrations (Appendix Table~\ref{tab:single_image}). The control average score over the five TrueMICL tasks reaches 61.5, above the untrained model (59.0) but far below using \data{} (76.2). Together, the two experiments show that stating the rule reduces the gain when the demonstrations are kept, and nearly eliminates it when they are removed.

\subsection{Training data analysis}
\label{sec:data_analysis}

\paragraph{Train-test separation.} To verify that the gains reflect generalization rather than memorization, we measure the distance between training and evaluation images. Each image is embedded by a frozen SigLIP~\cite{tschannen2025siglip} encoder and L2-normalized. We calibrate distances per source between two anchors: the \emph{floor} is the centroid distance to a held-out split of the same source, and the \emph{ceiling} is the distance to COCO photographs. A calibrated value near 0\% means the benchmark is visually indistinguishable from that training source; near 100\% means it is as distant as natural photographs.

Every evaluation benchmark lies at 70--170\% calibrated distance from its nearest training source, i.e., closer to unrelated photographs than to the training data. The one exception is CLEVR-based tasks (CLEVR-Color and VL-CLEVR). They sit at 8\% from CLEVR, with 93\% of their images above 0.95 cosine similarity to a training image, so we exclude them from evaluation. A t-SNE projection of the embeddings is in Appendix Figure~\ref{fig:train_src_tsne}.

\paragraph{\data{} properties} We ensure the data adhere to the properties listed in \S\ref{subsec::data}. We check all of the instances (Appendix~\ref{app:design_choices}).
\noindent\textbf{Text alone.} With the images removed, a text-only Qwen3-8B reaches near chance on average.

\noindent\textbf{Instance disjointness.} No query repeats a demonstration image or its underlying instance (0 violations).

\subsection{\method{} builds a stable, usable visual pathway}
\label{subsec:visual_pathway}

\method{} achieves comparable accuracy to fine-tuning on most \vicl{} tasks (\S\ref{subsec:res_dtad}). Yet, \method{} builds a stable visual pathway that is robust across tasks, and shot counts. Specifically, it absorbs the largest share of attention when demonstration text is blocked (Table~\ref{tab:attn_redistribution}).

\begin{figure*}[t]
    \centering
    \resizebox{\linewidth}{!}{%
    \begin{tabular}{ccc}
        \includegraphics[width=0.32\linewidth]{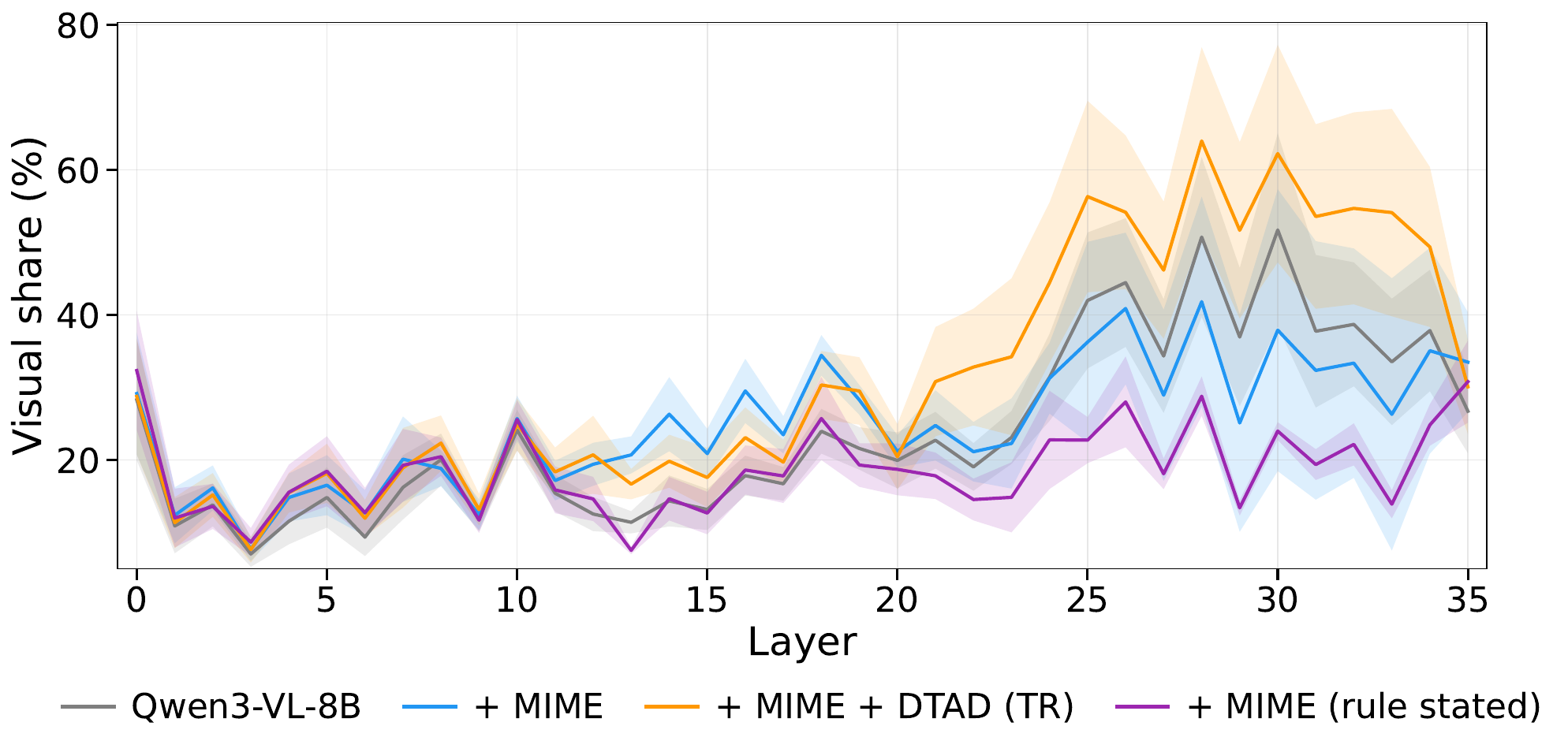} &
        \includegraphics[width=0.32\linewidth]{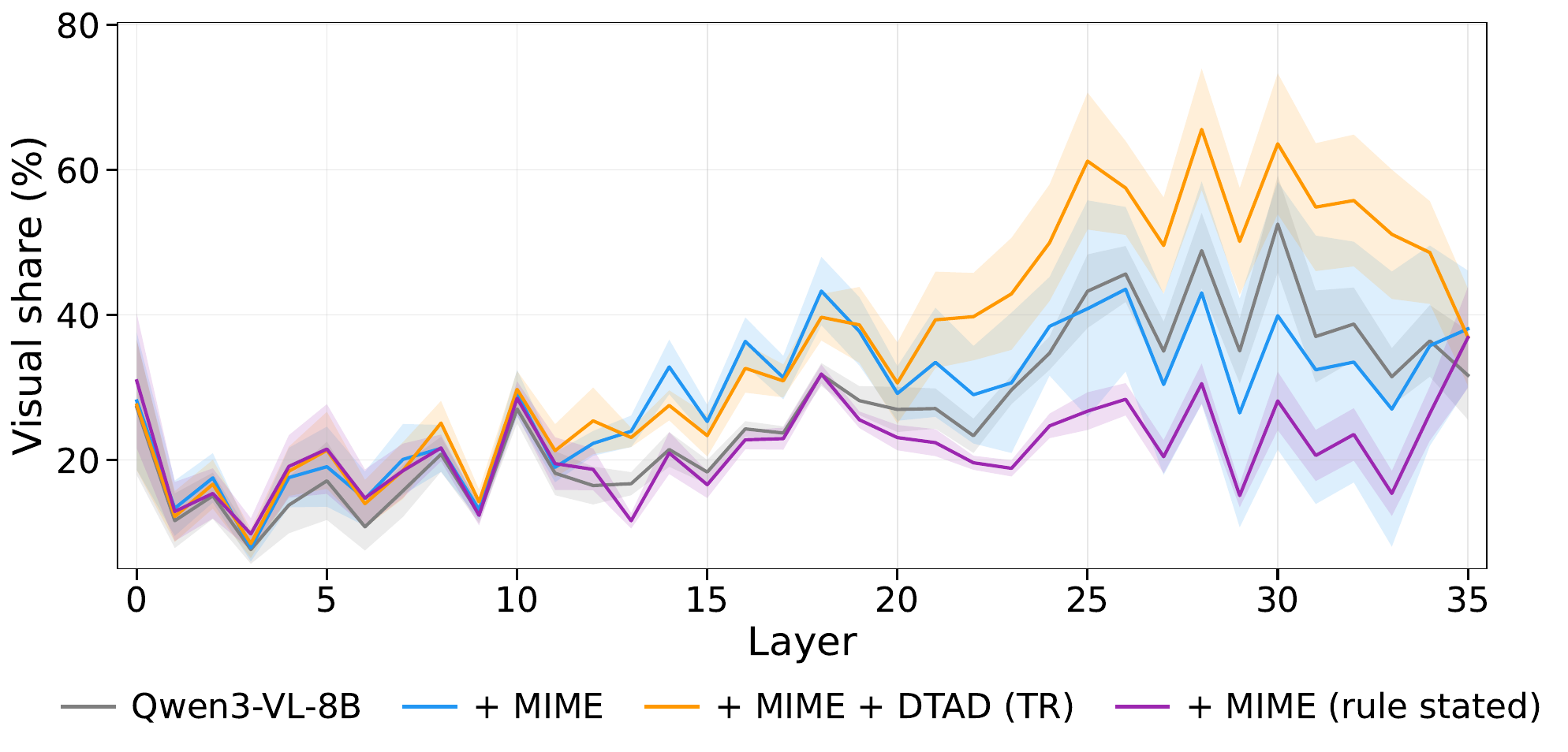} &
        \includegraphics[width=0.32\linewidth]{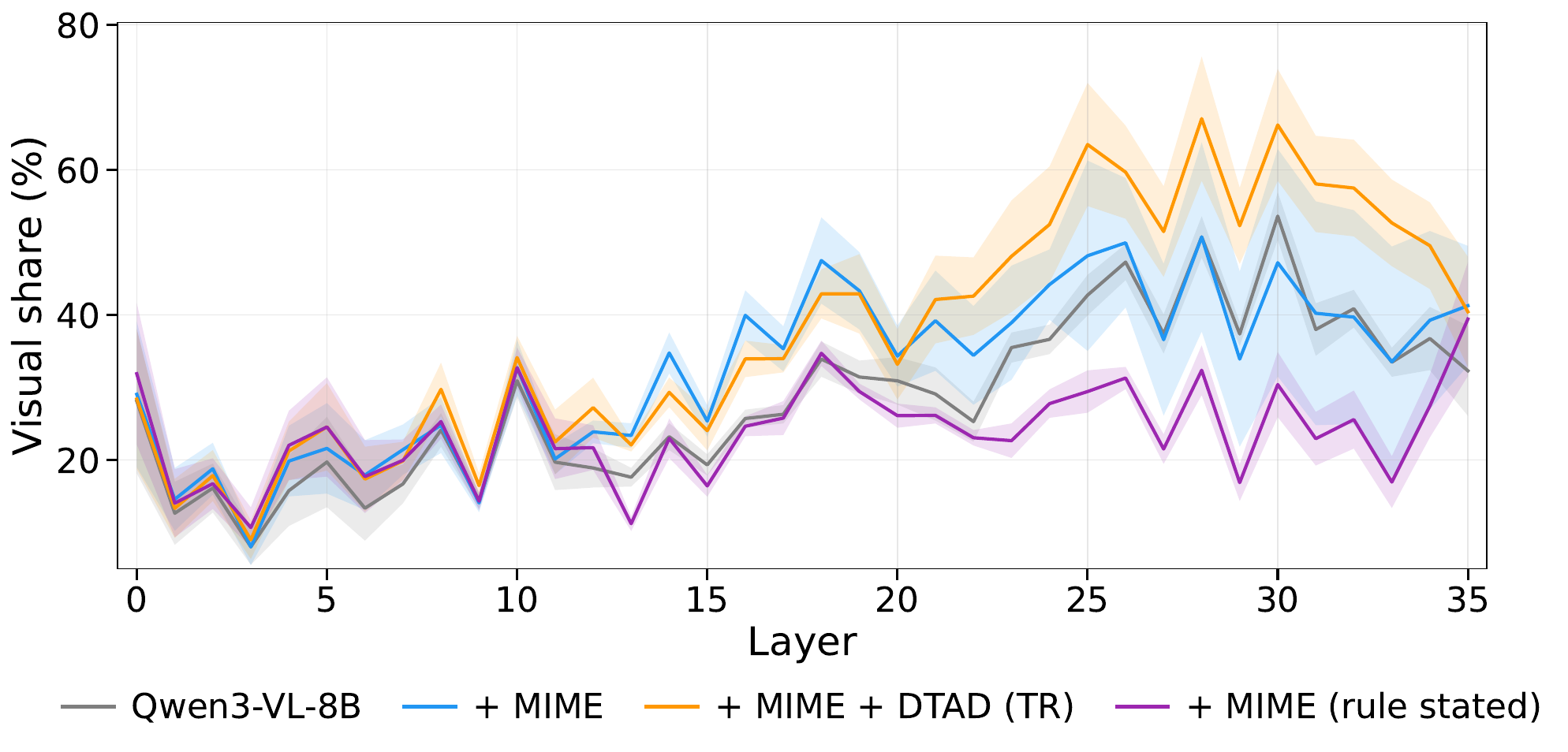} \\
    \end{tabular}}
    \vspace{-10pt}
    \caption{Mean visual share per layer for 1-shot (left), 2-shot (center), and 4-shot (right); lines show the mean over the 5 TrueMICL tasks and shaded bands show $\pm1$ SD across tasks. \method{} raises the visual share in the late layers above the base model at every shot count, with a narrow spread across tasks. Fine-tuning on the same data increases it mainly in the middle layers; in the late layers it varies widely across tasks. Training with the rule stated in the prompt lowers the visual share below the base model, indicating that the model relies mostly on the textual pathway.}
    \label{fig:visual_heads}
    \vspace{-6pt}
\end{figure*}

\paragraph{\method{} raises visual attention in late layers.}

\begin{equation}
  s_h = \frac{\sum_{q\in\mathcal{Q}}\sum_{t\in\mathcal{I}} a_{h,q,t}}
             {\sum_{q\in\mathcal{Q}}\sum_{t\in\mathcal{I}} a_{h,q,t} + \sum_{q\in\mathcal{Q}}\sum_{t\in\mathcal{T}} a_{h,q,t}},
  \label{eq:visual_share}
\end{equation}

For each head $h$, we take the attention from the query text tokens $\mathcal{Q}$ to the demonstrations. The visual share $s_h$ (Eq.~\ref{eq:visual_share}) is the fraction of this attention that goes to the demonstration image tokens $\mathcal{I}$ rather than to the demonstration text tokens $\mathcal{T}$.
For each layer, we average $s_h$ over its heads, and we report the mean and standard deviation over the 5 TrueMICL tasks. Figure~\ref{fig:visual_heads} shows the visual share of each layer at 1, 2, and 4 shots. In layers 0--11, all models overlap. The differences appear only in the middle and late layers, consistent with prior works~\citep{neo2025towards,cohen2025performance, kaduri2025s}.
Fine-tuning increases the visual share mainly in the middle layers (by 6--9 pp over the base model in layers 12--23). In the late layers (24--35), it has the largest spread across tasks (12--14 pp SD). When the prompt states the rule, the model relies on the text: its visual share is below the base model on mid-to-late layers. The untrained model also has a low visual share. This shows that pretraining alone does not make the model attend to visual demonstrations. Thus, dedicated \vicl{} training data is necessary.
\method{} raises the visual share above the base model from the middle layers onward, and the gap widens in the late layers, where \method{} is above the base model at every layer and every shot count. Its late-layer visual share stays nearly constant across shot counts, and its spread across tasks decreases as the number of shots increases.
Prior work shows that models extract object information in mid-to-late layers~\citep{neo2025towards,cohen2025performance, kaduri2025s} and also that heads which encode fine-grained visual details are in the deepest layers~\citep{chen2025ict}. Our results agree with these findings, and \method{} strengthens the late-layer visual pathway that already exists in the base model, and makes it consistent across shot counts and tasks.

\paragraph{Attention redistribution.} We block attention to all demonstration-text positions at inference and measure where the freed attention goes (Table~\ref{tab:attn_redistribution}). For all models, as the number of shots increases, the share that moves to the demonstration image grows.
\method{} absorbs the largest share at every shot count, reaching 31.3\% at 4 shots, compared with 24.6\% for the untrained model. Fine-tuning follows closely (29.7\% at 4 shots) but has the largest spread across tasks at every shot count, while training with the rule stated stays close to the untrained model (20.2--26.9\%). \method{} also performs consistently across tasks. It exceeds the untrained model in 14 of the 15 task--shot pairs (12 of 15 for fine-tuning).
Together, these results show that \data{} teaches the model to redirect attention to the demonstration images, and that \method{} pushes this further, with the largest and most consistent redirection across tasks.

\begin{table}[t]
\centering
\small
\caption{\textbf{Attention redistribution under demonstration-text blocking.} Absorption: share (\%) of the attention freed by blocking demonstration text that moves to demonstration images, all layers, mean $\pm$ SD over 5 TrueMICL tasks.}
\label{tab:attn_redistribution}
\vspace{-6pt}
\resizebox{\columnwidth}{!}{%
\begin{tabular}{lccc}
\toprule
Model & 1 shot & 2 shots & 4 shots \\
\midrule
Qwen3-VL-8B & 19.9 $\pm$ 2.0 & 20.6 $\pm$ 1.7 & 24.6 $\pm$ 1.8 \\
+ \data{} (rule stated) & 20.2 $\pm$ 3.2 & 22.3 $\pm$ 2.5 & 26.9 $\pm$ 2.9 \\
+ \data{} & 22.1 $\pm$ 3.4 & 26.1 $\pm$ 4.8 & 29.7 $\pm$ 3.5 \\
+ \data{} + \method{} (TR) & \textbf{22.9} $\pm$ 2.6 & \textbf{26.3} $\pm$ 2.8 & \textbf{31.3} $\pm$ 3.3 \\
\bottomrule
\end{tabular}}
\vspace{-10pt}
\end{table}

\vspace{-5pt}
\subsection{Scaling with shot count}
\label{sec:shot_scaling}

To assess our approach's ability to generalize beyond the shot counts seen during training, we evaluate on five TrueMICL tasks with 4 and 16 shots. We report the error reduction rate (ERR), namely: $(\text{acc@16} - \text{acc@4}) / (100 - \text{acc@4})$.

Table~\ref{tab:shots_headroom} shows that \method{} obtains the largest ERR (69.0\%), followed by fine-tuning on \data{} (67.5\%), both above the untrained model (53.9\%). Although the trained models start from a much higher 4-shot accuracy (88.6 and 90.0 vs.\ 66.6), leaving less room to improve, both \method{} and fine-tuning improve as demonstrations are added, and relative to their remaining headroom, they improve more than the untrained model.

\begin{table}[t]
\centering
\small
\caption{\textbf{Error reduction from 4 to 16 shots (TrueMICL).} Error reduction is the fraction of the 4-shot error removed at 16 shots. \method{} and fine-tuning keep improving with more demonstrations. The higher ERR demonstrates they improve more relative to their stronger starting point.}
\label{tab:shots_headroom}
\vspace{-6pt}
\resizebox{\columnwidth}{!}{%
\begin{tabular}{lccc}
\toprule
 & Qwen3-VL-8B & + \data{} & + \method{} (TR) \\
\midrule
Acc@4                  & 66.6   & 88.6   & 90.0   \\
Acc@16                 & 84.6   & 96.3   & 96.9   \\
Error reduction        & 53.9\% & 67.5\% & 69.0\% \\
\bottomrule
\end{tabular}}
\vspace{-15pt}
\end{table}

\section{Related Work}
\vspace{-4pt}
\label{sec:rw}
\textbf{Visual ICL models and benchmarks.}
Few-shot prompting was a design goal of interleaved VLMs from the outset: Flamingo trains on interleaved image-text web sequences and reports large few-shot gains on captioning and VQA \citep{alayrac2022flamingo}, a recipe reproduced by OpenFlamingo \citep{awadalla2023openflamingo} and extended to multi-image instruction tuning by Otter~\citep{li2023otter}. Interleaved data is now a standard pretraining ingredient \citep{laurenccon2024matters}, included explicitly to enable in-context learning \citep{bai2025qwen3, an2025llava, team2025granite}, so few-shot ability is assumed rather than specially built for.
Few benchmarks evaluate visual ICL directly. VL-ICL Bench \citep{zong2024vlicl} argues that few-shot VQA and captioning do not test ICL's limits and instead covers perception, induction, reasoning, and long-context tasks; TrueMICL \citep{chen2025true} claims VLMs imitate demonstration text while neglecting visual cues, and builds tasks that require both modalities; SEED-Bench-2 \citep{li2024seed} includes in-context captioning from two image-caption pairs as one dimension of a broader interleaved suite.

\textbf{Addressing the visual ICL gap.}
Prior work traces the failure to attention allocation \citep{chen2025true}: VLMs attend to demonstration text over demonstration images, leaving the visual half of each shot largely unused. \citet{wang2026multimodal} instead locate it earlier, in how the task mapping is built and propagated across layers.
Three families of fixes follow. The first reweights attention toward demonstration images: DARA trains the first attention layer to attend to them \citep{chen2025true}, while \citet{wang2026multimodal} rebalance attention at inference. Both give only marginal gains over the untrained model.
The second steers activations instead: multimodal task and in-context vectors, the analogue of task and function vectors in LMs \citep{todd2024function}, are extracted from many-shot activations and injected at inference, one per task \citep{huang2024multimodal, li2025m2iv}.
The third leaves the model untouched and fits per-task demonstration selection policies \citep{zhang2023makes, doveh2024towards, doveh2025teaching}. Separately, \citet{jia2025symdpo} penalize text shortcutting via preference optimization against symbol-substituted demonstrations.
These methods are typically task- or domain-specific and do not yield a general capability.

\section{Conclusions}
\label{sec:conc}
Vision-language models learn less from visual demonstrations than from textual ones: they route attention to demonstration text rather than images. We introduced \data{}, a cross-task \vicl{} training set in which the rule is carried only by the demonstrations and never stated. Fine-tuning on it improves \vicl{} by 8.1--18.0 points across several architectures, while preserving single-image ability and improving multi-image and text-only ICL. Our analysis shows that trained models rely on their demonstrations, and that this dependence arises because the rule is hidden: stating it reduces the gain significantly. \method{}, which withholds demonstration text during training, performs on par when the rule must come from the demonstrations, while strengthening late-layer attention to the demonstration images, consistently across shot counts and tasks.
A controlled full retrain of LLaVA-1.5-13B further shows that \data{} outperforms same-size multi-image data, demonstrating its benefit for the general instruction tuning of VLMs. 
We hope these results will establish \vicl{} data and demonstration-text dropout as standard components of VLM training.
\clearpage

\section*{Limitations}

Our controlled instruction-tuning experiment is limited to LLaVA-1.5-13B, since the training data and pipelines of current state-of-the-art VLMs are not public. We therefore cannot measure the full effect of adding \data{} to the pretraining and instruction-tuning stages of modern VLMs. Still, adding \data{} to the LLaVA-1.5 mix more than doubles its \vicl{} accuracy, which suggests that modern training pipelines could benefit as well.

Our fine-tuning experiments are similarly limited in scale. We train 4B--8B models with LoRA adapters, and leave full-parameter training with larger backbones unexplored. Even this lightweight setup lets an 8B model match frontier models on TrueMICL, leaving clear room for further gains under stronger training regimes.

Finally, we evaluate \data{} and \method{} in isolation and do not combine them with other approaches to \vicl{}. Because both act on the training data and objective, they are complementary to inference-time methods such as demonstration retrieval and task vectors, and combining them offers a promising path to further gains.

\clearpage
\bibliography{custom}

\clearpage

\appendix
\section{Appendix}
\label{sec:appendix}
This appendix collects the full experimental tables and additional analyses referenced in the main text. We begin with the motivating experiment that isolates the visual ICL bottleneck (Table~\ref{tab:app_prelim}). We then ablate the prompt format along the multi-turn/single-turn and instruction/no-instruction axes (Table~\ref{tab:format_ablation}). Next, we detail the training set composition (Table~\ref{tab:dataset_composition}) and analyze the contribution of each data source via leave-one-out training (Table~\ref{tab:loo}). Finally, we report a full-model retraining experiment on LLaVA-1.5-13B, including POPE and MME regression checks and a multi-image QA control (Table~\ref{tab:llava_training}).

\subsection{Preliminary Study}
\label{app:prem_study}
To locate the source of the visual ICL gap, we evaluate Qwen3-VL-8B on TrueMICL Clock and Operator Induction tasks while progressively removing textual descriptions of the demonstrations. Table~\ref{tab:app_prelim} shows that the model solves the task perfectly once the relevant information is available as text, even when it extracts that information from the images itself, but fails when it must infer the task directly from visual demonstrations.

\begin{table}[H]
\centering
\small
\resizebox{\columnwidth}{!}{%
\begin{tabular}{llccc}
\toprule
Setting & Shots & Clock & Operator induction & Mean \\
\midrule
\multirow{4}{*}{Pure text-ICL}
 & 1 & 100.0 & 99.0  & 99.5  \\
 & 2 & 100.0 & 100.0 & 100.0 \\
 & 4 & 100.0 & 100.0 & 100.0 \\
 & \textit{Avg} & \textbf{100.0} & \textbf{99.7} & \textbf{99.8} \\
\midrule
\multirow{4}{*}{Transcribe \& answer}
 & 1 & 100.0 & 99.0  & 99.5  \\
 & 2 & 100.0 & 100.0 & 100.0 \\
 & 4 & 100.0 & 100.0 & 100.0 \\
 & \textit{Avg} & \textbf{100.0} & \textbf{99.7} & \textbf{99.8} \\
\midrule
\multirow{4}{*}{\vicl{}, task instruction (baseline)}
 & 1 & 56.3 & 77.3 & 66.8 \\
 & 2 & 69.7 & 82.3 & 76.0 \\
 & 4 & 85.3 & 98.7 & 92.0 \\
 & \textit{Avg} & \textbf{70.4} & \textbf{86.1} & \textbf{78.3} \\
\midrule
\multirow{4}{*}{\vicl{}, generic instruction}
 & 1 & 9.3  & 62.0 & 35.7 \\
 & 2 & 11.7 & 76.7 & 44.2 \\
 & 4 & 17.3 & 91.3 & 54.3 \\
 & \textit{Avg} & \textbf{12.8} & \textbf{76.7} & \textbf{44.8} \\
\bottomrule
\end{tabular}}
\caption{\textbf{Preliminary study: what limits \vicl{}?} Accuracy (\%) of Qwen3-VL-8B on two TrueMICL tasks (Clock, Operator induction). \emph{Pure text-ICL}: the task is posed as text-only ICL using the task metadata. \emph{Transcribe \& answer}: the model first transcribes the images, then answers. \emph{Task instruction (baseline)}: standard \vicl{} with the official TrueMICL prompt. \emph{Generic instruction}: \vicl{} with an instruction that gives no hint about the task. \textit{Avg}: mean over shots; Mean: mean over the two tasks. Near-perfect text-ICL and transcribe \& answer scores show that perception and textual ICL are not the bottleneck, while the drop under a generic instruction indicates reliance on textual cues.}
\label{tab:app_prelim}
\end{table}

\twocolumn[{%
\subsection{Results}
\label{app:results}

\subsubsection{Official protocol results}
\label{app:official}
Table~\ref{tab:app_official} reports per-task results under the official protocol, which keeps each benchmark's task-specific question. Fine-tuning on \data{} improves all three backbones under this protocol as well (+7.8, +9.7, and +31.6 points), while DARA and MGI remain within seed variation of the untrained model.

\par\vspace{4pt}
{\centering\footnotesize
\captionof{table}{\textbf{Visual ICL results under the official protocol.} Per-task accuracy (\%) with each benchmark's task-specific question, averaged over 1/2/4 shots and three demonstration seeds; subscripts give the standard deviation of the overall average across seeds.}
\label{tab:app_official}
\vspace{3pt}
\resizebox{\ifdim\width>\linewidth\linewidth\else\width\fi}{!}{%
\begin{tabular}{l|ccccc|c|ccccc|c}
\toprule
 & \multicolumn{5}{c|}{TrueMICL} & SB-2 & \multicolumn{5}{c|}{VL-ICL-Bench} & \\
Model & Clock & Op.Ind & Palin & Shapes & Classif & T23 & Op.Ind & Interl & TextOCR & Open MI & Match MI & Overall \\
\midrule
Qwen3-VL-8B                 & 70.4 & 86.1 & 96.4 & 81.0 & 50.0 & 35.8 & 78.1 & 66.9 & 57.2 & 60.3 & 85.8 & 69.8$_{\pm0.6}$ \\
\quad + DARA                & 70.3 & 85.6 & 96.6 & 80.3 & 49.5 & 35.8 & 78.5 & 66.9 & 57.5 & 60.5 & 85.8 & 69.8$_{\pm0.5}$ \\
\quad + MGI                 & 70.2 & 85.8 & 96.7 & 81.2 & 49.1 & 35.8 & 77.6 & 66.7 & 57.5 & 60.7 & 85.2 & 69.7$_{\pm0.4}$ \\
\quad + \data{}                & 86.4 & 85.7 & 96.7 & 92.0 & 52.8 & 50.0 & 86.3 & 80.2 & 55.4 & 83.1 & 85.3 & \textbf{77.6}$_{\pm0.2}$ \\
\quad + \data{} + \method{}    & 86.1 & 83.4 & 96.6 & 97.8 & 43.2 & 45.8 & 83.1 & 78.0 & 62.0 & 82.2 & 81.5 & 76.3$_{\pm0.7}$ \\
\midrule
LLaVA-OV-1.5-8B             & 51.2 & 78.4 & 76.2 & 85.8 & 34.4 & 41.7 & 76.7 & 26.9 & 38.9 &  9.2 & 54.7 & 52.2$_{\pm0.4}$ \\
\quad + \data{}                & 64.1 & 86.6 & 83.0 & 98.0 & 38.2 & 40.8 & 84.6 & 39.1 & 49.8 & 35.9 & 60.8 & \textbf{61.9}$_{\pm0.4}$ \\
\quad + \data{} + \method{}    & 48.6 & 80.2 & 84.6 & 98.6 & 41.1 & 34.2 & 85.7 & 48.1 & 49.6 & 17.1 & 57.5 & 58.7$_{\pm0.4}$ \\
\midrule
Granite-Vision-4.1-4B       & 13.0 &  3.1 & 41.2 & 26.7 & 21.7 & 30.8 &  0.6 &  0.4 & 25.8 & 24.1 & 50.0 & 21.6$_{\pm0.3}$ \\
\quad + \data{}                & 41.1 & 78.3 & 62.7 & 91.7 & 39.2 & 30.8 & 77.6 & 18.5 & 35.4 & 56.7 & 53.4 & \textbf{53.2}$_{\pm0.7}$ \\
\quad + \data{} + \method{}    & 29.1 & 67.6 & 82.4 & 94.6 & 38.9 & 29.2 & 70.9 & 22.0 & 29.7 & 16.8 & 50.1 & 48.3$_{\pm0.3}$ \\
\bottomrule
\end{tabular}}\par}
}]

\twocolumn[{%
\subsubsection{Prompt Format Ablation}
\label{app:format_ablation}
Our adapters are trained with a multi-turn prompt that includes a generic instruction. Table~\ref{tab:format_ablation} evaluates all models under the four combinations of turn structure and instruction presence to test robustness to this choice.

\par\vspace{4pt}
{\centering\footnotesize
\captionof{table}{\textbf{Prompt format ablation.} TrueMICL accuracy under four prompt formats: \{multi-turn, single-turn\} $\times$ \{with instruction, without instruction\}. Multi-turn with instruction is the format the adapters were trained on. \emph{Overall} is the unweighted mean of Clock, Op.Ind, Palindrome, Shapes and Classification (each averaged over 1/2/4 shots). $\Delta_\text{MT+I}$ is the difference from the multi-turn + instruction cell. Trained models score higher with multi-turn prompts, while the untrained baseline scores highest with single-turn + instruction (60.5). Removing the instruction costs the baseline and FT 6.8--7.2~pp on multi-turn; \method{} models are the most robust to it ($-$0.5 and $-$2.1~pp).}
\label{tab:format_ablation}
\vspace{3pt}
\resizebox{0.85\linewidth}{!}{%
\begin{tabular}{ll|ccccc|c|c}
\toprule
 & & \multicolumn{5}{c|}{TrueMICL} & & \\
\cmidrule(lr){3-7}
Format & Model & Clock & Op.Ind & Palin & Shapes & Classif & Overall & $\Delta_\text{MT+I}$ \\
\midrule
\multirow{4}{*}{\shortstack[l]{Multi-turn\\+ instruction}}
 & Qwen3-VL-8B (baseline) & 12.8 & 76.7 & 63.9 & 95.8 & 45.7 & 59.0 & - \\
 & \quad + \data{}             & 62.8 & 88.9 & 68.8 & 97.3 & 63.4 & 76.2 & - \\
 & \quad + \data{} + \method{} (Attn Mask) & 62.3 & 85.3 & 91.7 & 99.9 & 37.8 & 75.4 & - \\
 & \quad + \data{} + \method{} (Token Replace)  & 70.1 & 88.1 & 79.3 & 99.2 & 53.9 & 78.1 & - \\
\midrule
\multirow{4}{*}{\shortstack[l]{Multi-turn\\no instruction}}
 & Qwen3-VL-8B (baseline) & \phantom{0}6.7 & 82.0 & 44.0 & 92.6 & 33.7 & 51.8 & $-$7.2 \\
 & \quad + \data{}            & 48.3 & 89.7 & 67.0 & 96.1 & 45.9 & 69.4 & $-$6.8 \\
 & \quad + \data{} + \method{} (Attn Mask) & 62.0 & 84.7 & 83.3 & 99.6 & 37.0 & 73.3 & $-$2.1 \\
 & \quad + \data{} + \method{} (Token Replace)  & 64.7 & 94.0 & 86.0 & 99.2 & 44.3 & 77.6 & $-$0.5 \\
\midrule
\multirow{4}{*}{\shortstack[l]{Single-turn\\+ instruction}}
 & Qwen3-VL-8B (baseline) & \phantom{0}9.0 & 89.3 & 71.7 & 93.6 & 39.0 & 60.5 & $+$1.5 \\
 & \quad + \data{}             & 53.0 & 86.3 & 73.3 & 98.3 & 42.7 & 70.7 & $-$5.5 \\
 & \quad + \data{} + \method{} (Attn Mask) & 57.3 & 83.0 & 87.7 & 99.8 & 45.1 & 74.6 & $-$0.8 \\
 & \quad + \data{} + \method{} (Token Replace)  & 59.0 & 88.7 & 82.0 & 98.8 & 46.7 & 75.0 & $-$3.1 \\
\midrule
\multirow{4}{*}{\shortstack[l]{Single-turn\\no instruction}}
 & Qwen3-VL-8B (baseline) & \phantom{0}7.7 & 78.0 & 11.0 & 87.9 & 39.4 & 44.8 & $-$14.2 \\
 & \quad + \data{}             & 47.7 & 86.0 & 43.3 & 95.1 & 39.8 & 62.4 & $-$13.8 \\
 & \quad + \data{} + \method{} (Attn Mask) & 54.3 & 83.3 & 73.3 & 99.1 & 41.1 & 70.2 & $-$5.2 \\
 & \quad + \data{} + \method{} (Token Replace)  & 58.7 & 87.7 & 74.0 & 98.7 & 44.3 & 72.7 & $-$5.4 \\
\bottomrule
\end{tabular}}\par}
\vspace{4pt}
}]

\twocolumn[{%
\subsubsection{Beyond \vicl{} tasks}
\label{app:beyond}
Table~\ref{tab:app_beyond} reports per-benchmark results for \S\ref{subsec:res_beyond}. Fine-tuning on \data{} preserves single-image performance (76.1 vs.\ 76.4 on average), improves both multi-image benchmarks, and raises accuracy on MARVEL, as well as on all three text-only ICL benchmarks.

\par\vspace{4pt}
{\centering\footnotesize
\captionof{table}{\textbf{Beyond visual ICL: per-benchmark results.} Single-image and multi-image benchmarks check that training on \data{} does not regress general ability; MARVEL and text-only ICL test transfer. MARVEL is evaluated as multi-turn visual ICL (chance 25\%). Text-only ICL uses multi-turn demonstrations. MME is reported as perception/cognition.}
\label{tab:app_beyond}
\vspace{3pt}
\resizebox{\ifdim\width>\linewidth\linewidth\else\width\fi}{!}{%
\begin{tabular}{l|cccccc|ccc|c|ccc}
\toprule
 & \multicolumn{6}{c|}{Single-image} & \multicolumn{3}{c|}{Multi-image} & & \multicolumn{3}{c}{Text-only ICL} \\
Model & MMStar & MME P/C & RealWorldQA & POPE & InfoVQA & Avg & Mantis & LLaVA-Interleave & Avg & MARVEL & MMLU & GSM8K & GPQA \\
\midrule
Qwen3-VL-8B  & 63.6 & 1723.4/641.1 & 69.4 & 88.8 & 82.7 & 76.1 & 69.1 & 40.6 & 54.9 & 27.0 & 74.6 & 89.8 & 36.4 \\
\quad + \data{} & 65.1 & 1735.1/649.3 & 68.9 & 88.9 & 82.7 & 76.4 & 73.3 & 41.7 & 57.5 & 29.1 & 75.6 & 91.4 & 37.1 \\
\quad + \data{} + \method{} & 65.4 & 1749.4/637.9 & 71.4 & 88.5 & 82.4 & 76.9 & 71.4 & 44.7 & 58.1 & 31.6 & 75.8 & 92.7 & 37.1 \\
\bottomrule
\end{tabular}}\par}
\vspace{4pt}


}]

\subsection{Data}
\label{app::data}

\subsubsection{Statistics}
\label{app:data_stats}
Our training set combines five sources, each contributing up to 5{,}000 episodes with 1--5 demonstrations per episode. Table~\ref{tab:dataset_composition} lists the per-source episode counts and shot distributions.

\begin{table}[H]
\centering
\caption{Composition of the training set by source, showing the number of episodes, the range of shots used, and the mean number of shots per episode.}
\label{tab:dataset_composition}
\vspace{3pt}
\resizebox{\ifdim\width>\linewidth\linewidth\else\width\fi}{!}{%
\begin{tabular}{lrcr}
\toprule
Source & Episodes & Shots & Mean shots \\
\midrule
PuzzleVQA polygon sides        & 5000  & $\{2,3,4\}$   & 2.94 \\
Bongard-LOGO shape concept     & 5000  & $\{2,4\}$     & 3.00 \\
RE-ARC transform induction     & 5000  & $\{3,4,5\}$   & 3.97 \\
PuzzleVQA triangle arithmetic  & 5000  & $\{2,3,4\}$   & 3.00 \\
CLEVR                          & 3578  & $\{1,2,4\}$   & 2.64 \\
\midrule
Total                          & 23578 & $\{1,\dots,5\}$ & 3.14 \\
\bottomrule
\end{tabular}}
\end{table}

\subsubsection{Single-Image Control}
\label{app:single_image}
To separate the effect of the demonstration structure from exposure to the training images, we train a control on the same query images with each episode's rule stated in text and no demonstrations, using the same hyperparameters. Table~\ref{tab:single_image} shows that the control improves over the untrained model (61.5 vs.\ 59.0) but stays well below training on our \vicl{} data (76.2). The gap is sharpest on Clock, whose rule exists only in the visual demonstrations: the control does not improve over the untrained model (11.0 vs.\ 12.8), while \method{} reaches 70.1. The control's gain comes from Palindrome (+16.4) and Operator Induction (+12.0), whose rules are easily stated in words, and it drops below the untrained model on Classification ($-$12.4) and Shapes ($-$1.5).

\begin{table}[t]
\centering
\caption{\textbf{Single-image control.} TrueMICL per-task accuracy (\%, Qwen3-VL-8B, averaged over 1/2/4 shots). The control trains on the same query images with each rule stated in text and no demonstrations. It improves over the untrained model but falls far short of training on our \vicl{} data. The gap is largest on Clock, whose rule exists only in the visual demonstrations.}
\label{tab:single_image}
\vspace{3pt}
\resizebox{\ifdim\width>\linewidth\linewidth\else\width\fi}{!}{%
\begin{tabular}{lcccccc}
\toprule
Model & Clock & Op.Ind & Palin & Shapes & Classif & Avg \\
\midrule
Qwen3-VL-8B                 & 12.8 & 76.7 & 63.9 & 95.8 & 45.7 & 59.0 \\
Single-image control        & 11.0 & 88.7 & 80.3 & 94.3 & 33.3 & 61.5 \\
\midrule
+ \data{}                      & 62.8 & 88.9 & 68.8 & \textbf{97.3} & 63.4 & \textbf{76.2} \\
\bottomrule
\end{tabular}}
\end{table}

\subsubsection{Dataset design choices}
\label{app:design_choices}
We verify each property of our dataset directly on the 23,578 training episodes. The demonstration-only and hidden-rule properties are tested by corrupting demonstrations and by retraining with the rule stated (Tables~\ref{tab:corrupt} and~\ref{tab:rule_stated}); Table~\ref{tab:source_audit} reports, per source, text-only predictability, 
and query--demonstration instance violations.

\begin{table}[t]
\centering
\small
\setlength{\tabcolsep}{4pt}
\caption{\textbf{Per-source audit (\%).} Text-only predictability and query-demonstration instance violations.}
\label{tab:source_audit}
\vspace{3pt}
\resizebox{\ifdim\width>\linewidth\linewidth\else\width\fi}{!}{%
\begin{tabular}{lccccc}
\toprule
 & Polygon & Bongard & RE-ARC & Triangle & CLEVR \\
\midrule
Chance                              & 7.1 & 50.0 & 14.3 & 4.5 & 4.5 \\
Qwen3-8B text-only                  & 23.0 & 27.5 & 4.5 & 11.5 & 4.5 \\
Query-demo violations (count)       & 0 & 0 & 0 & 0 & 0 \\
\bottomrule
\end{tabular}}
\end{table}

\twocolumn[{%
\subsubsection{Train-Test Separation}
\label{app:train_test}
Figure~\ref{fig:train_src_tsne} shows a t-SNE projection of the SigLIP~\cite{tschannen2025siglip} image embeddings used in \S\ref{sec:data_analysis}. Each training source forms a tight, well-separated cluster, and the evaluation benchmarks fall outside these clusters, at 70--170\% calibrated distance from their nearest training source (0\% = a held-out split of that source, 100\% = COCO photographs). The exception is CLEVR-based benchmarks, which lie within the CLEVR training cluster (8\% calibrated distance, with 93\% of its images above 0.95 cosine similarity to a training image); we therefore exclude them from evaluation.

\par\vspace{6pt}
{\centering
\includegraphics[width=\linewidth,height=0.45\textheight,keepaspectratio]{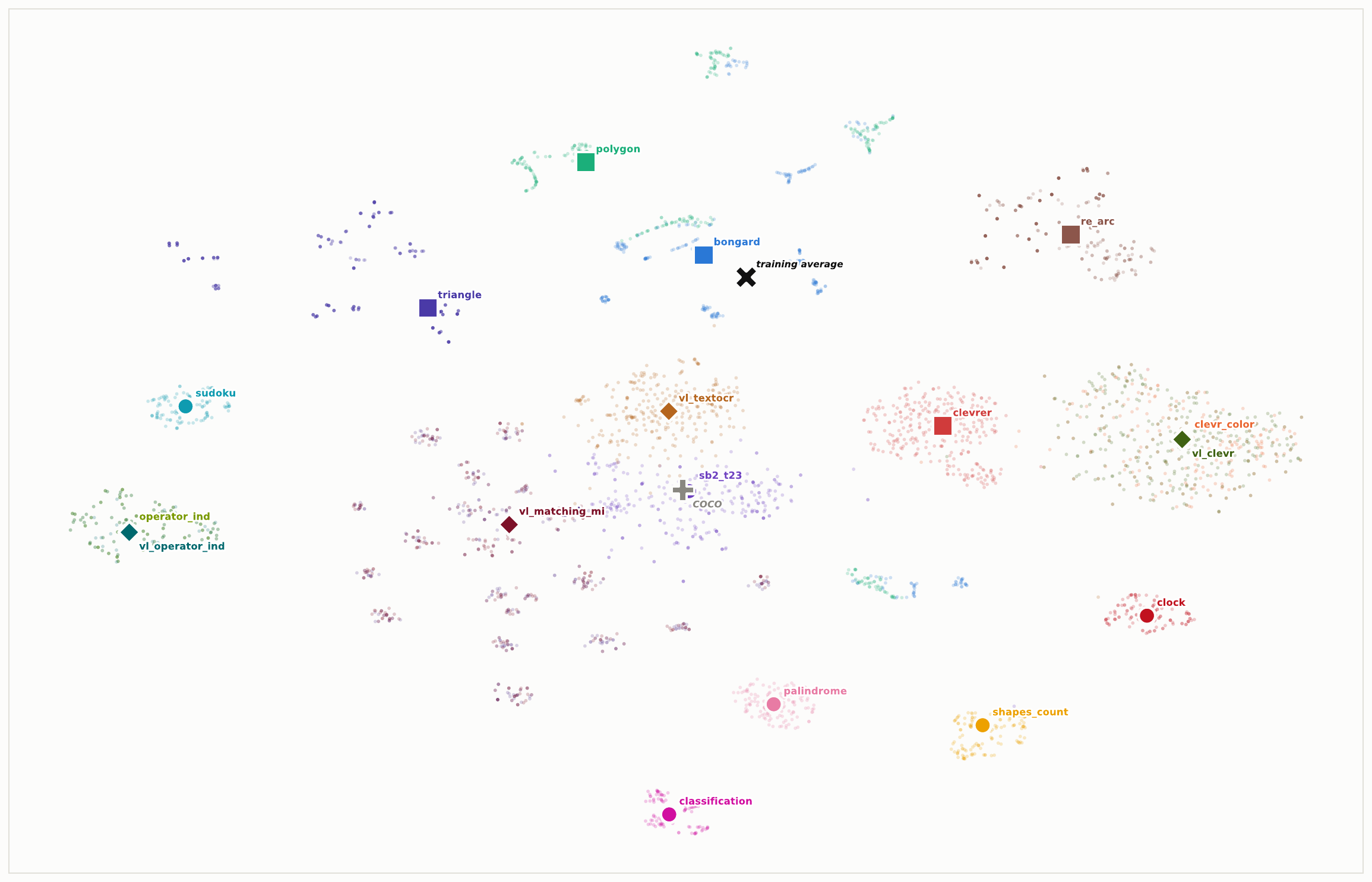}
\captionof{figure}{t-SNE of SigLIP image embeddings. Training sources (squares) form tight clusters. Evaluation benchmarks (circles, diamonds) lie at 70--170\% calibrated distance from the nearest training source (0\% = held-out split of that source, 100\% = COCO), except CLEVR-based benchmarks (8\% to CLEVR), which we therefore exclude from evaluation.}
\label{fig:train_src_tsne}\par}
\vspace{4pt}
}]

\subsubsection{Source Contribution}
\label{app:source_contribution}
To measure the contribution of each source, we train leave-one-out (LOO) adapters, each omitting one source, under all three training strategies, evaluating each at 1, 2, and 4~shots on the TrueMICL benchmarks (Table~\ref{tab:loo}).

The value of source diversity depends on the training strategy. Under both \method{} variants, removing any source costs 3.0--16.7 points, whereas under naïve fine-tuning only removing Triangle has a clear effect ($-$10.2); the other changes are within $\pm$2.6 points. CLEVR is the clearest case: removing it costs 16.7 points under \method{} (AM) but only 1.0 under naïve fine-tuning.

\begin{table}[H]
\centering
\small
\caption{\textbf{Leave-one-out.} Change in TrueMICL average accuracy when one source is removed from the full mix.}
\label{tab:loo}
\vspace{3pt}
\resizebox{\ifdim\width>\linewidth\linewidth\else\width\fi}{!}{%
\begin{tabular}{lccc}
\toprule
Removed source & Naïve FT & \method{} (AM) & \method{} (TR) \\
\midrule
Polygon      & $-$2.6 & $-$3.0 & $-$9.7 \\
Bongard-LOGO & $+$2.6 & $-$3.9 & $-$6.5 \\
RE-ARC       & $-$2.4 & $-$4.1 & $-$4.6 \\
Triangle     & $-$10.2 & $-$8.3 & $-$7.4 \\
CLEVR        & $-$1.0 & $-$16.7 & $-$6.7 \\
\bottomrule
\end{tabular}}
\end{table}

\twocolumn[{%
\subsection{Full-Model Retraining on LLaVA-1.5-13B}
\label{app:llava_training}
Finally, we test whether \data{} also helps when mixed into a model's original instruction-tuning stage: we retrain LLaVA-1.5-13B from its 665K instruction mix with our visual ICL data added, and compare against the official checkpoint and a multi-image QA control trained on Mantis (Table~\ref{tab:llava_training}).

\par\vspace{4pt}
{\centering\footnotesize
\captionof{table}{\textbf{Full-model retraining: LLaVA-1.5-13B with visual ICL data.} TrueMICL mean accuracy for the official LLaVA-1.5-13B model, a model retrained with our 5-source visual ICL data mixed into the llava training data (\data), and a multi-image (MI) QA control (Mantis, 23.5K episodes, no ICL pattern). \data{} training gives large ICL gains (+26 pp Clock, +32 pp Op.Ind, +46 pp Shapes; +20.8 pp average) with no regression on POPE or MME. The Mantis control shows that multi-image exposure alone does not teach visual ICL (+2.6 pp average), and it causes a severe POPE regression ($-$30 pp, yes-bias).}
\label{tab:llava_training}
\vspace{3pt}
\resizebox{\ifdim\width>\linewidth\linewidth\else\width\fi}{!}{%
\begin{tabular}{l ccccc c ccc}
\toprule
 & \multicolumn{6}{c}{TrueMICL (avg 1/2/4-shot)} & \multicolumn{3}{c}{Regression check} \\
\cmidrule(lr){2-7} \cmidrule(lr){8-10}
Model & Clock & Op.Ind & Palin & Shapes & Classif & Avg & POPE & MME-P & MME-C \\
\midrule
Official & 5.3 & 5.0 & 25.7 & 35.9 & 22.6 & 18.9 & 84.2 & 1561 & 118 \\
MI       & 5.7 & 9.0 & \textbf{30.3} & 41.0 & 21.4 & 21.5 & 54.1 & 1509 & 116 \\
\data{}     & \textbf{31.7} & \textbf{37.3} & 25.3 & \textbf{81.9} & 22.2 & \textbf{39.7} & \textbf{85.2} & \textbf{1575} & \textbf{127} \\
\bottomrule
\end{tabular}}\par}
\vspace{4pt}
}]

\end{document}